\documentclass[preprint,12pt,authoryear]{elsarticle}

\usepackage{amsmath,amssymb,mathtools}
\usepackage{booktabs}
\usepackage{float}
\usepackage{microtype}
\usepackage{graphicx}
\usepackage{tikz}
\usetikzlibrary{arrows.meta,positioning,calc}
\usepackage[hidelinks]{hyperref}

\journal{Cognitive Systems Research}

\begin{document}

\begin{frontmatter}

\title{
On Representational Alignment among Embodied Agents
}

\author{
Fulvio Mastrogiovanni\\
University of Genoa, Italy
}

\begin{abstract}
Embodied agents interacting with the same physical process may maintain heterogeneous, asynchronous, and observer-relative representations.
Rather than assuming that such representations should always be globally aligned, we investigate which distinctions among them must actually be resolved for coherent interaction.
We formalize this question through \emph{relational sufficiency}: an interaction-specific relational-evidence map defines the representational ambiguity left unresolved by comparison, and sufficiency holds when each residual ambiguity fiber lies within an interaction-equivalence class.
When this condition fails, additional \emph{sensorimotor anchors} must separate precisely the ambiguities that can change the interaction-relevant outcome.
For \textit{smooth} systems, we derive a local first-order lower bound on the number of independent scalar anchors required, given by the dimension of directions that are invisible to relational evidence but visible to the interaction-relevant map.
We instantiate the theory in two variants of an asynchronous hand-off scenario that share the same observer agents, representations, transports, and relational evidence.
Relative synchronization requires no additional anchor, whereas hand-off at an absolute time requires exactly one.
The analysis therefore provides a task-relative stopping condition for representational alignment: residual disagreement need not be eliminated when it is invariant with respect to the interaction.
\end{abstract}


\end{frontmatter}

\section{Introduction}
\label{sec:introduction}

Embodied, cognitive agents do not observe and act in the world from a common viewpoint. 
Their access to an environment is mediated by their body morphologies, sensors, action capabilities, interaction histories, and internal
representational processes. 
Even when two agents cooperate by interacting with the same physical process, the representations available to them may differ in reference frame, time, content, abstraction level, or semantic form. 
Partial and asynchronous observations further amplify these differences. 
From the perspective of embodied cognitive systems, this raises a basic question: what kind of representational compatibility is actually required for coherent interaction between two or more embodied agents?

A common approach is to seek some form of representational \textit{alignment}.
Different methods have been pursued toward this aim.
Distributed estimation methods aim at deriving mutually compatible estimates of a shared state;
multi-agent systems exchange information to coordinate private beliefs;
multimodal and cross-agent representation-learning methods construct mappings between heterogeneous latent spaces;
recent work has shown that local comparison spaces can support alignment even when a single global latent space is absent \citep{Robinson2017, Moschella2023, Ghalkha2026, DAcunto2026}.
Similarly, recent embodied multi-agent architectures explicitly study the alignment of private world models through communication \citep{Dongre2026}.
These developments substantially \textit{weaken} the requirement that different agents should maintain identical internal representations. 
They nevertheless retain, in different forms, an objective of making the representations relevant to coordination mutually compatible.

This paper investigates a different possibility. 
We ask whether some representational disagreements can remain unresolved because they are irrelevant to the interaction task. 
More precisely, we consider embodied observer agents whose local representations may be heterogeneous and not directly comparable; 
when required, observer-specific representations are \textit{transported} into an interaction-dependent comparison context; 
the resulting joint relational configuration may still be compatible with multiple underlying representational configurations. 
We refer to this residual indeterminacy as \emph{relational ambiguity}. 
Our central hypothesis is that such ambiguity need not be eliminated \textit{in general}. 
It only needs to be resolved along those representational \textit{degrees of freedom} that can change the task-relevant outcome.

This small shift changes the question from how two representations should be globally aligned to which \emph{relations between them} must be determined.
The distinction is especially important in the case of embodied agents. 
For such agents, representations are not \textit{detached} descriptions, as they participate in a closed sensorimotor loop in which sensing produces observer-relative information, representations support action, actions modify subsequent interaction conditions, and new observations constrain what the agents can continue to represent. 
Therefore, the environment may constrain local representations through the consequences of sensorimotor interaction without providing the agents with a
privileged, globally accessible representation of itself.

A minimal example illustrates the issue. 
Let us consider two embodied observers coordinating around the future occurrence of an event, for example, to determine when such an event will occur.
The first observer represents the event through an estimated distance and velocity, while the second represents it directly through an estimated time-to-event. 
Their native representations differ, but both can be transported into estimates of event time. 
Let us suppose that the only relational evidence available after transport is the difference between the two estimates. 
This evidence is invariant to a common temporal shift of both estimates. 
If the interaction requires only relative synchronization, the common shift is therefore irrelevant, and every representation pair related by that shift supports the same task response. 
If, instead, the interaction requires both agents to act relative to an external physical event, the same ambiguity becomes consequential. 
Relative agreement is then insufficient, and some additional sensorimotor relation to the environment is required. 
It is noteworthy that the observers, their representations, their transport mechanisms, and their relational evidence can remain unchanged across the two cases, since only the task determines whether the unresolved ambiguity matters.

We use this contrast to develop a theory of \emph{relational sufficiency among embodied observer agents}. 
Let a joint configuration of transported representations be denoted by $q$, and let $C_I(q)$ denote the relational evidence available for an interaction $I$.
The configurations that cannot be distinguished through this evidence form an ambiguity class $\Omega_C(q)$. 
Independently, the interaction induces an equivalence relation over representational configurations, such that two configurations are equivalent when they cannot be distinguished by the task-relevant response. 
The resulting class is denoted by $[q]_\Theta$. 
Therefore, the basic sufficiency condition studied in this paper is

\begin{equation*}
\Omega_C(q) \subseteq [q]_\Theta.
\end{equation*}

It can be noted that this relation expresses a \textit{deliberately weaker} requirement than representational identification or global alignment.
Relational evidence is sufficient whenever all configurations that remain epistemically indistinguishable are also indistinguishable to the
interaction. 
The interacting agents do not have to determine which member of the equivalence class is the ``correct'' representation.

However, when this inclusion relation fails, relational evidence leaves unresolved a distinction that the interaction can detect. 
We introduce \emph{sensorimotor anchors} to characterize the additional information required in this case. 
An anchor is not a global \textit{world representation} or an \textit{oracle} supplying ground truth. 
On the contrary, it is an interaction-derived constraint that is available through the agents' admissible sensorimotor coupling.
For example, it could be an environmental cue, a contact event, a landmark observation, or other locally obtainable consequence of interaction. 
Anchors are sufficient when they separate those relational ambiguities that correspond to different task-relevant outcomes. 
This leads to a principle that guides our model, that is, an embodied agent should resolve only those relational ambiguities that the interaction itself can distinguish.

In this paper, we develop this principle formally and examine its consequences in two analytically solvable variants of the same hand-off scenario.
The two variants share the same observer agents, local representations, representational transports, and relational evidence, but differ in their task invariances. 
The first introduces an observer-relative synchronization task, in which the ambiguity preserved by relational evidence is entirely task-invariant and no additional anchor is required.
In the second, we discuss an absolute hand-off task, for which the same ambiguity is task-relevant and exactly one independent scalar
anchor is required. 
The comparison therefore isolates task structure, rather than representational accuracy or sensing capability, as the factor that determines how much relational alignment is necessary.

The main contributions of the paper are as follows.

\begin{enumerate}
\item 
We formulate \emph{relational sufficiency} for two embodied observer agents with heterogeneous local representations, and we explicitly distinguish relational epistemic evidence, task response, and interaction adequacy.
\item 
We establish a mismatch result showing that relational evidence is insufficient whenever it preserves ambiguities that the task can distinguish, and we characterize sufficient sensorimotor anchoring as the separation of such task-conflicting ambiguities.
\item 
We derive a local first-order lower bound on the number of independent scalar anchors required for relational sufficiency; the bound is given by the dimensional mismatch between directions that are invisible to relational evidence and directions that are invariant to the task.
\item 
We instantiate the theory in two asynchronous hand-off toy scenarios and show analytically that the same relational ambiguity requires no anchors for relative synchronization but one anchor for absolute interaction; we further show that adequacy can require less information than exact task-response reconstruction.
\item 
We characterize representational alignment as \emph{task-over-refined} whenever it distinguishes configurations lying within the same interaction-equivalence class; this provides a precise sense in which global alignment may be sufficient while being stronger than necessary for embodied interaction.
\end{enumerate}


The remainder of the paper first positions the problem with respect to existing work on representation alignment, sufficient representations, decentralized information, and task-relevant observability, while introducing the hand-off scenarios that serve as a running example.
Then, we define the sensorimotor and representational model of embodied observer agents, develop the relational sufficiency and anchoring results, and instantiate them analytically in the two hand-off scenarios. 
Finally, we discuss implications for representation alignment, learning for embodied agents, the degenerate single-observer case, and extensions to interactions involving more than two observer agents.

\section{Background and Motivating Scenario}
\label{sec:background}

\subsection{Shared representations and relational sufficiency}

The problem addressed in this paper lies at the intersection of several well-established research directions. 
Distributed estimation and sensor fusion have long considered systems in which multiple sensing units access different, noisy, and possibly delayed observations of a common dynamical process. 
Such methods provide mature mechanisms for combining heterogeneous measurements and handling partial or asynchronous information \citep{Capitan2011}. 
In robotics, problems like multi-robot simultaneous localisation and mapping similarly demonstrate that consistency can often be assessed through relative constraints rather than through direct access to an external ground truth. 
Pairwise consistency methods and their higher-order extensions, for example, identify mutually compatible inter-robot measurements through graphs or hypergraphs of relative constraints \citep{Mangelson2018, Forsgren2024}.
Estimation and SLAM also make explicit the role of unobservable directions: some transformations of the estimated state can leave the available relative measurements unchanged, and consistent estimation must avoid acquiring spurious information along those directions.
Our setting shares this concern with unresolved degrees of freedom, but asks an additional question: whether a direction that is invisible to the available relational evidence is also invisible to the interaction.

These approaches already establish an important principle for the work discussed here, that is, useful consistency relations can be defined locally and relationally.
However, their usual objective remains the recovery of a mutually compatible description, typically expressed in a geometric, dynamical, or symbolic form. 
The \textit{latent} object to be estimated or aligned may be decentralized, partially observed, or represented in multiple coordinate systems, but the different information sources are ultimately
interpreted with respect to a \textit{common} estimation problem.

Recent representation-learning methods weaken this requirement further.
Observer-relative representations allow independently learned latent spaces to be compared through relations to shared reference samples, which enables communication across otherwise incompatible latent coordinates
\citep{Moschella2023}. 
Sheaf-theoretic approaches provide a more general ontology for heterogeneous information sources.
Local data may remain in different spaces, while restriction or transport maps specify how information
is compared on overlaps \citep{Robinson2017}. 
Recent decentralized representation-learning methods extend this idea to multiple local comparison spaces and learned mappings, thus relaxing the requirement for a single shared latent space \citep{Ghalkha2026, DAcunto2026}.

A related strand of work appears in multi-agent world modelling. 
Embodied agents may maintain private and evolving internal descriptions of their environment, communicate selectively, and reason about what other agents know. 
Recent work has explicitly investigated the suitability of \textit{dialogues} as a mechanism for aligning such private world models under partial observability \citep{Dongre2026}. 
These approaches are particularly relevant because they hypothesise that representational disagreement is no longer merely a mismatch grounded in geometric, dynamical, or symbolic information.
On the contrary, it can involve different observation modalities, beliefs, semantic structures, and histories of interaction.

A second set of antecedents concerns \emph{sufficiency} rather than global alignment.
Decentralized stochastic control has long studied how agents with asymmetric information can act optimally from limited private and common information.
The common-information approach reformulates certain decentralized problems through information shared by the controllers \citep{Nayyar2013}, while the sufficient-information approach explicitly seeks mutually consistent
compressions that preserve decision quality \citep{Tavafoghi2018}. 
These results are close in principle to our work because they show that decision making need not necessarily retain all available information. 
Their starting point, however, is a common stochastic decision model coupled with agent-specific information structures. 
Our object of study is different, in that we ask which task-relevant discrepancies between heterogeneous observer-relative representations must become distinguishable at all.

Related ideas appear in state abstraction and representation learning for decision making. 
Bisimulation-based abstractions and approximate state abstractions identify states that can be merged without materially changing optimal behaviour \citep{Ferns2004, Abel2016}.
Learned latent-state models make a similar distinction between high-dimensional observations and compact representations sufficient for reward prediction and dynamics
\citep{Gelada2019}. 
These works establish that task-relevant \textit{quotienting}, that is, collapsing distinctions that do not matter for decision making, is not itself new. 
In our work, the quotient is instead defined over a \emph{joint relational configuration} assembled from representations that may originate in different observers, times, or representational spaces.

A particularly important related field of inquiry is functional observability.
Functional observers seek to reconstruct only a target function or a subset of a system state when reconstructing the complete state is unnecessary \citep{Montanari2022PNAS, Montanari2022PRR}. 
Recent work also studies minimum sensor placement for functional observability \citep{Zhang2025}. 
The distinction pursued in this work with respect to this literature is that the unobservable object is not simply a part of a single underlying state.
Instead, it is a \textit{degree of freedom} in the relation among heterogeneous representations, and the additional measurements are constrained to arise as \textit{admissible} sensorimotor anchors.

Finally, goal- and task-oriented communication explicitly prioritizes the information that matters to shared objectives rather than maximizing message fidelity \citep{Charalambous2026}. 
Also in this case, our question is adjacent but not identical. 
We do not optimize what agents communicate. 
We characterize which \emph{inter-agent ambiguities} must be eliminated, regardless of whether the relevant evidence is obtained through communication, sensing, or physical interaction.

In a general sense, the present work starts from these developments but poses a different question.
We do not investigate how heterogeneous representations maintained by different observer agents can always be made globally consistent, nor do we propose another mechanism for mapping them into a common latent space. 
Instead, we ask whether representational consistency should itself be treated as a task-dependent requirement.

This distinction is relevant because an embodied agent does not acquire, maintain, or use a representation in isolation. 
Representations participate in sensorimotor loops.
Observations are conditioned by the agent's embodiment and sensing capabilities; 
internal representations support action; 
actions modify the subsequent interaction with the environment;
the environment produces new observations and constraints. 
Classical active-perception accounts already emphasize that perception is inseparable from the actions through which information is acquired \citep{Bajcsy1988}. 
More recent embodiment-oriented accounts of cognition similarly stress that predictive internal models acquire their meaning through their role in
closed-loop sensorimotor interaction \citep{Pezzulo2024}.

Therefore, we deliberately adopt a \textit{relational} perspective. 
Two embodied observer agents may maintain different internal representations and may even remain globally incompatible, while still possessing enough relational structure to interact successfully. 
We posit that what matters is not necessarily whether their representations can be fully reconciled, but whether the distinctions that remain unresolved can alter the interaction.

This leads to three levels that we keep separate throughout the paper.
First, \emph{agreement} describes a relation between already comparable representational quantities, such as whether two estimates differ by less than
a specified tolerance. 
Second, \emph{epistemic congruity} refers more generally to whether the available relational evidence satisfies the constraints relevant to a particular interaction. 
Third, \emph{interaction adequacy} concerns the physical or operational outcome generated when those representations are used for action. 
Agreement can contribute to congruity, and congruity can contribute to successful interaction, but none of these notions is identified with the others.
This separation also distinguishes the problem we address from approaches whose goal is to maximize alignment. 
A residual disagreement may be harmless if the task is invariant to it. 
Conversely, even perfect agreement may be insufficient if both agents share an error along a factor that matters to the interaction. 
The theoretical problem is therefore not to minimize representational discrepancy indiscriminately, but to determine which relational ambiguities are consequential.

\subsection{An example scenario}

We introduce a simple scenario that will be analysed in depth later.
Its purpose at this stage is not to establish any result, but to make the distinction between agreement, relational sufficiency, and physical adequacy concrete.
Consider a one-dimensional environment containing an object moving at constant velocity $x(t)=x_0+vt$, with $v>0$, toward a hand-off location $x_H$.
The corresponding physical hand-off time is $t_H=(x_H-x_0)/v$.
This quantity belongs to the external specification of the scenario: it is available to an \textit{external observer} for defining and evaluating the interaction, but it is not assumed to be directly available to either observer agent.

Two observer agents access the process differently.
Agent $A$, observing at time $t_A$, represents the future hand-off through an estimated remaining distance and velocity, such that $R_A=(\hat d_A,\hat v_A)$.
Agent $B$, observing at a generally different time $t_B$, represents the same future event directly through an estimated time-to-hand-off, that is,
$R_B=\widehat{\tau}_B$.
Thus $R_A\in\mathcal R_A$ and $R_B\in\mathcal R_B$, with no requirement that $\mathcal R_A=\mathcal R_B$.

For the hand-off interaction, however, both representations can be transported into a comparison quantity corresponding to estimated absolute hand-off time:
\[
q_A=\Phi_A^{c_H}(R_A)=t_A+\frac{\hat d_A}{\hat v_A},
\qquad
q_B=\Phi_B^{c_H}(R_B)=t_B+\widehat{\tau}_B.
\]
The comparison context $c_H$ is interaction-specific.
It does not constitute a complete common world model; it merely specifies the quantity with respect to which the two representation events become comparable for this interaction.

Let the available relational evidence be their signed difference, $d=q_A-q_B$.
A simple agreement criterion is then $|d|\leq\epsilon$.
Crucially, this evidence is invariant under the common shift
\[
(q_A,q_B)\mapsto(q_A+\delta,q_B+\delta).
\]
The observer agents cannot distinguish configurations related by this shift from their pairwise difference alone.
Agreement therefore constrains their relative estimate while leaving a one-dimensional common-mode ambiguity unresolved.
Whether that ambiguity matters depends entirely on the interaction.

We have two variants of the same scenario.
In the first variant, denoted \textsf{S1-R}, the agents need only achieve \emph{relative synchronization}, for example $|q_A-q_B|\leq\eta_{AB}$.
The common shift leaves this requirement unchanged, so resolving the absolute temporal offset provides no additional information relevant to this interaction.
In the second variant, denoted \textsf{S1-A}, the agents must act with respect to the physical hand-off event itself.
A prototypical adequacy requirement is
\[
|q_A-t_H|\leq\eta_A,
\qquad
|q_B-t_H|\leq\eta_B.
\]
The same common shift now changes the observer agents' relation to the physical event while preserving their mutual agreement.
Two agents may therefore agree perfectly and nevertheless both fail the hand-off.

The contrast between \textsf{S1-R} and \textsf{S1-A} is deliberately minimal.
The physical process, observer agents, sensing models, native representations, representational transports, and pairwise evidence can remain identical.
Only the invariance structure of the interaction changes.
In \textsf{S1-R}, common-mode ambiguity is invisible both to the relational evidence and to the interaction; in \textsf{S1-A}, it is invisible to the relational evidence but visible to the interaction.

This pair of scenarios isolates the central theoretical issue of the paper: \emph{Does the ambiguity left unresolved by relational evidence contain distinctions that the interaction can detect?}
If the answer is negative, further representational alignment is unnecessary.
Otherwise, some additional relation to the environment must remove the interaction-relevant part of that ambiguity.
We call such additional relations \emph{sensorimotor anchors}.
An anchor does not provide a complete or privileged representation of the environment; it needs only supply enough interaction-derived information to distinguish configurations that are relationally indistinguishable but behaviourally different.

The next Section develops the formal model required to make these statements precise and falsifiable.
We first define embodied observer agents, representation events, and representational transport, and then distinguish relational evidence from task response and interaction adequacy.
This makes it possible to state relational ambiguity, interaction equivalence, and admissible sensorimotor anchoring without presupposing a globally shared representation.

\section{A Relational Model of Embodied Observer Agents}
\label{sec:model}

In this Section, we introduce the minimal formal model used throughout the paper.
The objective is not to specify a complete cognitive architecture, but to isolate the structures required to study relational sufficiency among embodied agents: embodiment, sensorimotor coupling, observer-relative representation, representational transport, and interaction-dependent comparison.

\subsection{Embodied observer agents and representation events}

We consider two embodied observer agents, indexed by $i$ and $j$, with $i\neq j$.
Each observer agent has a constitution $\Xi_i=(B_i,S_i,A_i)$, where $B_i$ denotes its embodiment, $S_i$ its sensing capabilities, and $A_i$ the set of actions available to it.
These components determine how the agent can couple to the environment, which observations can become available, and which interventions can alter future observations.

Let $x_t$ denote the physical state specifying the interaction dynamics.
An observer agent $i$ does not, in general, have direct access to $x_t$.
Instead, it receives a local observation according to
\[
o_i^t\sim\mathsf O_i(\cdot\mid x_t,\Xi_i),
\]
and accumulates a sensorimotor history $H_i^t=(o_i^{0:t},a_i^{0:t-1})$.
Different observer agents may have different embodiments, sensing modalities, sampling times, noise characteristics, and action spaces; even when they interact with the same physical process, their epistemic access need not coincide.

In the paper, we use the term \emph{observer} for an entity whose epistemic access is local and conditioned by $\Xi_i$, and \emph{agent} for an observer that can act so as to modify its subsequent coupling to the environment.
The expression \emph{embodied observer agent} emphasizes that sensing and acting participate in a closed sensorimotor loop.

From its sensorimotor history, the observer $i$ may construct a representation $R_\alpha=\rho_i(H_i^{t_\alpha},\kappa_\alpha)$ for a purpose, query, viewpoint, modality, or interaction context $\kappa_\alpha$.
We call the situated occurrence of such a representation at time $t_\alpha$ a \emph{representation event} $\alpha$.
For two distinct observers, we generally have $R_\alpha\in\mathcal R_i$ and $R_\beta\in\mathcal R_j$, with no assumption that $\mathcal R_i=\mathcal R_j$.
The representations may differ not only numerically but structurally, for example by encoding geometric or temporal quantities, symbolic relations, or learned latent variables.
Equality of local representations is therefore neither assumed nor generally meaningful.

Observer $i$ closes its sensorimotor loop through an action policy $a_i^t=\pi_i(H_i^t,R_\alpha,g_i^t)$, with $a_i^t\in A_i$ and interaction objective $g_i^t$.
Joint actions modify the interaction dynamics, for example through $x_{t+1}\sim\mathsf F_I(\cdot\mid x_t,a_i^t,a_j^t)$, thereby generating new observations.
The relevant causal structure is thus
\[
x_t\rightarrow o_i^t\rightarrow R_\alpha\rightarrow a_i^t \rightarrow x_{t+1}\rightarrow o_i^{t+1}.
\]
The use of $x_t$ does not imply that $x_t$ itself, or any representation derived from it, is available to the observer agents as a privileged shared world representation.

\subsection{Representational transport and relational evidence}

Because local representations may belong to different spaces, their comparison is not assumed to be immediately possible.
For an interaction $I$, let $c$ denote a \emph{comparison context}, which expresses the conditions under which a set of representation events can be made meaningfully comparable.
Depending on the interaction, $c$ may specify a reference time, spatial frame, task variable, semantic convention, or another comparison requirement.

For a representation event $\alpha$, a representational transport
\[
\Phi_\alpha^c:\mathcal R_i\rightarrow\mathcal Q_c
\]
produces a comparison quantity $q_\alpha=\Phi_\alpha^c(R_\alpha)$.
The comparison space $\mathcal Q_c$ is local to the comparison context.
It need not be a complete representation of the environment, nor need the same comparison space be used for other interactions.
Transport may correspond, for example, to temporal prediction, retrodiction, frame transformation, perspective taking, semantic translation, or the identity map when the representations are already comparable.

For two observer agents, we write the transported joint configuration as $q=(q_i,q_j)\in\mathcal Q_I$, where $\mathcal Q_I\subseteq\mathcal Q_c\times\mathcal Q_c$ is the joint configuration space relevant to interaction $I$.
This is a theoretical object used to characterize relations among transported representation events; the observers need not possess $q$ as centralized knowledge.

The available \emph{relational evidence} is represented by a map
\[
C_I:\mathcal Q_I\rightarrow\mathcal Z_I,
\]
where $C_I(q)$ contains those properties of the joint configuration that can be determined through the comparison mechanisms available in the interaction.
These may include relative differences, agreement residuals, ordering relations, semantic constraints, or communication outcomes.
An epistemic congruity criterion can then be evaluated as a predicate over $C_I(q)$, but the theory supports a full relational-evidence evaluation rather than only a Boolean congruity judgement.
Most importantly, relational evidence is specified independently of whether subsequent sensorimotor interaction succeeds.

\noindent\textit{Running example.}
For the hand-off scenario of Section~\ref{sec:background}, let $q=(q_A,q_B)$ and take $C_I(q)=q_A-q_B$.
Then, for example, $q=(5,3)$ and $q'=(6,4)$ produce the same evidence, $C_I(q)=C_I(q')=2$.
The relational comparison therefore cannot distinguish these configurations, even though their absolute temporal locations differ.
This simple common-shift ambiguity will serve as a running example throughout the theoretical development.

\subsection{Task response, relational ambiguity, and sensorimotor anchors}

Relational evidence describes what can be established representationally; it does not yet specify what the interaction itself distinguishes.
We therefore introduce a task-response map
\[
\Psi_I:\mathcal Q_I\rightarrow\mathcal Y_I.
\]
For a fixed interaction instance and fixed action policy, $\Psi_I(q)$ denotes the task-relevant physical or behavioural response associated with $q$.
We assume that $q$ contains the transported variables on which that response depends; otherwise those variables must be included before $\Psi_I$ is well defined as a function of $q$.

Interaction adequacy is evaluated separately through a map $J_I:\mathcal Y_I\rightarrow\mathcal P_I$, giving $\mathcal A_I(q)=J_I(\Psi_I(q))$.
For a binary notion of success, $\mathcal P_I=\{0,1\}$.
Thus relational evidence, task response, and interaction adequacy are distinct levels of description: agreement belongs to the first, the physical consequence of acting from the representations to the second, and success or failure to the third.

The ambiguity left unresolved by relational evidence is the fiber of $C_I$ through $q$, that is
\[
\Omega_C(q)
=
C_I^{-1}(C_I(q))
=
\{q'\in\mathcal Q_I:C_I(q')=C_I(q)\}.
\]
To express which distinctions matter for the interaction, let $\Theta_I:\mathcal Q_I\rightarrow\mathcal Y_\Theta$ denote an \emph{interaction-relevant map}.
It induces the equivalence relation $q\sim_\Theta q'$ whenever $\Theta_I(q)=\Theta_I(q')$, with equivalence class
\[
[q]_\Theta
=
\{q'\in\mathcal Q_I:\Theta_I(q')=\Theta_I(q)\}.
\]
Two choices will be useful: $\Theta_I=\Psi_I$ gives \emph{response equivalence}, whereas $\Theta_I=\mathcal A_I$ gives the weaker notion of \emph{adequacy equivalence}.
An exact reconstruction of the task response can therefore require more information than is needed to guarantee an adequate interaction.

\noindent\textit{Running example.}
With $C_I(q)=q_A-q_B$, the configurations $q=(5,3)$ and $q'=(6,4)$ lie in the same ambiguity fiber.
If the interaction-relevant map is the relative response $\Theta_R(q)=q_A-q_B$, they also lie in the same interaction-equivalence class.
If instead $\Theta_A$ depends on absolute timing with respect to the physical hand-off event, the same common shift can move the two configurations into different interaction-equivalence classes.
The relational evidence is unchanged; what changes is whether the interaction can distinguish the unresolved ambiguity.

If relational evidence leaves unresolved configurations belonging to different interaction-equivalence classes, additional information is required.
We restrict such information to \emph{admissible sensorimotor anchors}.
In the deterministic setting considered here, an anchor is a locally obtainable interaction-derived quantity $\zeta_k:\mathcal Q_I\rightarrow\mathcal Y_k$ whose value can be acquired through the sensing or interaction capabilities available to the observer agents before the interaction-related decision or action is completed.
Examples include detecting an environmental landmark, observing a contact event, receiving an acknowledgement, or acquiring a measurement through an admissible sensing action.

For a set of anchors $K$, let us write $\zeta_K(q)=(\zeta_k(q))_{k\in K}$ and let us define the augmented evidence as
\[
C_{I,K}(q)=\big(C_I(q),\zeta_K(q)\big).
\]
Its residual ambiguity is the corresponding fiber
\[
\Omega_{C,K}(q)
=
\{q':C_{I,K}(q')=C_{I,K}(q)\}.
\]
An anchor is therefore not a declaration of ground truth and need not identify a unique global configuration.
Its role is only to remove ambiguity that cannot safely remain unresolved for the interaction.

The central sufficiency condition studied in the remainder of the paper is
\begin{equation}
\Omega_{C,K}(q)\subseteq[q]_\Theta.
\label{eq:relational_sufficiency_condition}
\end{equation}
After relational comparison and any necessary sensorimotor anchoring, every configuration that remains indistinguishable must also be equivalent with respect to the chosen interaction-relevant map.
Crucially, Equation~\eqref{eq:relational_sufficiency_condition} does not require $\Omega_{C,K}(q)=\{q\}$: residual representational ambiguity is permitted whenever the interaction cannot distinguish it.

\begin{figure}[t!]
\centering
\begin{tikzpicture}[
    scale=0.82,
    transform shape,
    node distance=7mm and 11mm,
    box/.style={draw, rounded corners, align=center, minimum height=8mm,
                text width=31mm},
    small/.style={draw, rounded corners, align=center, minimum height=7mm,
                  text width=29mm},
    arr/.style={-{Latex[length=2mm]}, thick}
]

\node[box] (oa) {Embodied observer $i$\\$\Xi_i,H_i$};
\node[box, right=of oa] (ob) {Embodied observer $j$\\$\Xi_j,H_j$};

\node[small, below=of oa] (ra) {Representation event\\$\alpha:R_\alpha$};
\node[small, below=of ob] (rb) {Representation event\\$\beta:R_\beta$};

\node[small, below=of ra] (ta) {Transport\\$\Phi_\alpha^c$};
\node[small, below=of rb] (tb) {Transport\\$\Phi_\beta^c$};

\node[box, below=12mm of $(ta)!0.5!(tb)$] (q)
{Joint comparison configuration\\$q=(q_i,q_j)$};

\node[box, below left=9mm and 8mm of q] (evidence)
{Relational evidence\\$C_I(q)$};

\node[box, below right=9mm and 8mm of q] (interaction)
{Sensorimotor interaction\\and environmental cues};

\node[box, below=of interaction] (anchors)
{Admissible anchors\\$\zeta_K(q)$};

\node[box, below=14mm of $(evidence)!0.5!(anchors)$] (aug)
{Augmented evidence\\$C_{I,K}(q)$};

\node[box, below=of aug] (target)
{Relational sufficiency\\$\Omega_{C,K}(q)\subseteq[q]_\Theta$};

\draw[arr] (oa) -- (ra);
\draw[arr] (ob) -- (rb);
\draw[arr] (ra) -- (ta);
\draw[arr] (rb) -- (tb);
\draw[arr] (ta) to[bend right=12] (q);
\draw[arr] (tb) to[bend left=12] (q);
\draw[arr] (q) to[bend right=12] (evidence);
\draw[arr] (q) to[bend left=12] (interaction);
\draw[arr] (interaction) -- (anchors);
\draw[arr] (evidence) to[bend right=12] (aug);
\draw[arr] (anchors) to[bend left=12] (aug);
\draw[arr] (aug) -- (target);

\end{tikzpicture}
\caption{
Conceptual architecture of relational sufficiency.
Observer-relative representations become comparable through interaction-specific transport.
Relational evidence may leave residual ambiguity.
Sensorimotor interaction supplies admissible anchors only when that ambiguity crosses interaction-equivalence classes.}
\label{fig:relational_architecture}
\end{figure}
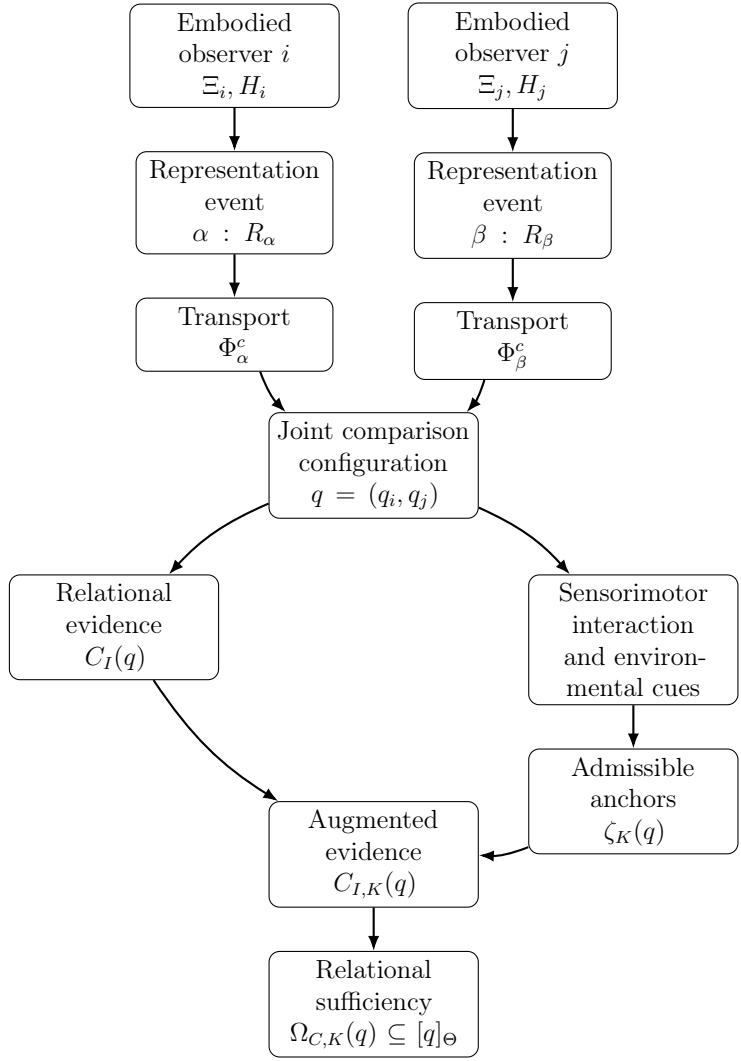

Interaction consequences may subsequently generate new observations and representation events, closing the observer agents' sensorimotor loops.
Figure~\ref{fig:relational_architecture} summarizes the distinction between observer-relative representation, constructed comparability, relational evidence, and interaction-derived anchoring.

The next Section studies when Equation~\eqref{eq:relational_sufficiency_condition} holds, when additional anchoring is necessary, and how much sensorimotor anchoring is minimally required.

\section{Relational Sufficiency and Minimal Anchoring}
\label{sec:theory}

Section~\ref{sec:model} separated relational evidence $C_I(q)$, task response $\Psi_I(q)$, and interaction adequacy $\mathcal A_I(q)$.
We now ask when the relational information available to embodied observer agents contains everything relevant to a specified interaction despite leaving part of the joint representational configuration unresolved.
Throughout this Section, $\Theta_I$ denotes the interaction-relevant map introduced above.
As we mentioned, choosing $\Theta_I=\Psi_I$ gives \emph{response sufficiency}; choosing $\Theta_I=\mathcal A_I$ gives \emph{adequacy sufficiency}.

\subsection{Relational sufficiency and interaction invariance}

The first result states the exact condition under which relational evidence already contains all information needed for the chosen interaction-relevant quantity.

\noindent\textit{Proposition 1 (Relational sufficiency)}
The following statements are equivalent:
\begin{enumerate}
\item[(i)] there exists a map $f_\Theta:\operatorname{Im}(C_I)\rightarrow\mathcal Y_\Theta$ such that $\Theta_I=f_\Theta\circ C_I$;
\item[(ii)] $C_I(q)=C_I(q')$ implies $\Theta_I(q)=\Theta_I(q')$;
\item[(iii)] $\Omega_C(q)\subseteq[q]_\Theta$ for every $q\in\mathcal Q_I$.
\end{enumerate}
When these conditions hold, $C_I$ is said to be \emph{$\Theta$-sufficient}.

\noindent\textit{Proof.}
(i)$\;\Rightarrow\;$(ii) follows by applying $f_\Theta$ to equal evidence values.
(ii)$\;\Leftrightarrow\;$(iii) is immediate from the definitions of the ambiguity fiber and the interaction-equivalence class.
For (iii)$\;\Rightarrow\;$(i), define $f_\Theta(z)=\Theta_I(q)$ for any $q$ satisfying $C_I(q)=z$.
Condition (iii) guarantees that all representatives of the same fiber have the same value of $\Theta_I$, so $f_\Theta$ is well defined.
\hfill$\square$

Proposition~1 formalizes the distinction from identification-oriented formulations, that is, residual ambiguity is not an error \textit{per se}.
The observer agents need not identify a unique configuration $q$.
Instead, they need only enough information to determine its interaction-equivalence class.

\noindent\textit{Running example.}
Take again $C_I(q_A,q_B)=q_A-q_B$.
The points $q=(5,3)$ and $q'=(6,4)$ belong to the same fiber because both give evidence $2$.
For the relative interaction map $\Theta_R(q_A,q_B)=q_A-q_B$, they also have the same interaction-relevant quantity.
Hence this common-shift ambiguity is harmless and $C_I$ is $\Theta_R$-sufficient.

\noindent\textit{Corollary 1 (Relational mismatch).}
If there exist $q,q'\in\mathcal Q_I$ such that $C_I(q)=C_I(q')$ but $\Theta_I(q)\neq\Theta_I(q')$, then $C_I$ is not $\Theta$-sufficient.
Equivalently, if $\Omega_C(q)\nsubseteq[q]_\Theta$ for some $q$, relational evidence is insufficient for $\Theta_I$.

\noindent\textit{Proof.}
Such a pair contradicts condition (ii) of Proposition~1.
\hfill$\square$

For $\Theta_I=\mathcal A_I$, Corollary~1 has a direct behavioural interpretation.
If identical relational evidence is compatible with both adequate and inadequate interaction, no decision rule based on that evidence alone can distinguish the two cases.

\noindent\textit{Running example.}
Let us suppose instead that the interaction-relevant map depends on absolute timing with respect to the physical hand-off event, for example
$\Theta_A(q)=(q_A-t_H,q_B-t_H)$.
The common shift $(q_A,q_B)\mapsto(q_A+\delta,q_B+\delta)$ preserves $C_I=q_A-q_B$ but changes $\Theta_A$ whenever $\delta\neq0$.
Thus the same ambiguity that is harmless for relative synchronization produces a relational mismatch for the absolute hand-off.

The same result can be expressed in terms of invariances.
Let us suppose that a family of transformations $G_C$ acts on $\mathcal Q_I$ and generates exactly the ambiguity fibers of the relational evidence, so that $\Omega_C(q)=G_C\cdot q$.
Let
\[
G_\Theta
=
\{g:\Theta_I(gq)=\Theta_I(q)\ \text{for all }q\in\mathcal Q_I\}
\]
denote the transformations invisible to the interaction-relevant map.
When the relevant transformation families form groups, Proposition~1 yields the following criterion.

\noindent\textit{Corollary 2 (Interaction-invariance criterion).}
If $\Omega_C(q)=G_C\cdot q$, then $C_I$ is $\Theta$-sufficient if and only if $G_C\subseteq G_\Theta$.
Thus every transformation invisible to relational evidence must also be invisible to the interaction.
Whenever $G_C\nsubseteq G_\Theta$, at least one representational degree of freedom is epistemically unresolved but interaction-relevant.

\textit{Running example.}
$G_C$ is the family of common translations $g_\delta(q_A,q_B)=(q_A+\delta,q_B+\delta)$.
For relative synchronization these transformations also belong to $G_{\Theta_R}$, whereas for absolute hand-off they do not.
The mismatch is therefore between the symmetries of the relational evidence and those of the interaction.

The same perspective identifies the amount of representational resolution fundamentally required.
The quotient map $\kappa_\Theta(q)=[q]_\Theta$ collapses all configurations that are equivalent for the chosen interaction-relevant map.
Any $\Theta$-sufficient representation $E:\mathcal Q_I\rightarrow\mathcal Z$ must determine this quotient, that is, there exists a map $h$ such that $\kappa_\Theta=h\circ E$.
The proof is immediate because $E(q)=E(q')$ implies $q\sim_\Theta q'$.
Thus $\mathcal Q_I/{\sim_\Theta}$ is the coarsest exact representation that preserves the chosen interaction-relevant quantity.

We call an aligned representation $L:\mathcal Q_I\rightarrow\mathcal U$ \emph{task-over-refined} whenever it distinguishes configurations within the same interaction-equivalence class, that is, whenever $q\sim_\Theta q'$ but $L(q)\neq L(q')$ for some pair $q,q'$.
Such an alignment may still be useful or sufficient, but it preserves distinctions that cannot alter $\Theta_I$ and is therefore stronger than required for this interaction.

\subsection{Minimal sensorimotor anchoring}

When relational evidence is not $\Theta$-sufficient, the observer agents require additional information.
As discussed in Section~\ref{sec:model}, such information is restricted here to admissible sensorimotor anchors.
Their role is not to recover a unique global representation, but to remove precisely those ambiguities that prevent relational sufficiency.

Let us define the set of \emph{conflicting epistemic pairs}
\[
\mathcal D_\Theta
=
\{(q,q'):C_I(q)=C_I(q'),\ \Theta_I(q)\neq\Theta_I(q')\}.
\]
These are pairs that relational evidence cannot distinguish but the interaction-relevant map can.

\noindent\textit{Proposition 2 (Admissible anchor separation).}
An anchor set $K$ makes the augmented evidence $C_{I,K}$ $\Theta$-sufficient if and only if every pair in $\mathcal D_\Theta$ is separated by at least one anchor:
\[
\forall(q,q')\in\mathcal D_\Theta,
\qquad
\exists k\in K:\zeta_k(q)\neq\zeta_k(q').
\]

\noindent\textit{Proof.}
If some conflicting pair is not separated, then it has identical relational evidence and identical anchor values, hence identical augmented evidence, while its $\Theta_I$ values differ; by Corollary~1, $C_{I,K}$ cannot be sufficient.
Conversely, if every conflicting pair is separated, equality of augmented evidence rules out membership in $\mathcal D_\Theta$, so equal augmented evidence implies equal $\Theta_I$ values.
Proposition~1 then gives sufficiency.
\hfill$\square$

Proposition~2 makes the role of interaction-derived grounding explicit.
An anchor need not identify the world state or reconstruct a world model; it need only separate ambiguities that the interaction can distinguish.

\noindent\textit{Running example.}
For the first variant we consider, \textsf{S1-R}, common-shift pairs are not conflicting because the relative interaction map is invariant to them, so no anchor is needed along that direction.
For \textsf{S1-A}, the same pairs are conflicting.
A sensorimotor cue carrying one independent piece of absolute temporal information can separate them without reconstructing the complete physical state.

For a finite or discretized problem, minimum-cost anchoring has a simple conflict-cover form.
Let $S_k\subseteq\mathcal D_\Theta$ be the set of conflicting pairs separated by the anchor $k$, and let $c_k\geq0$ be its sensing, communication, or interaction cost.
Then a minimum-cost sufficient set solves
\[
K^\star
=
\arg\min_K\sum_{k\in K}c_k
\qquad
\text{subject to}
\qquad
\bigcup_{k\in K}S_k=\mathcal D_\Theta.
\]
It can be observed that the objective is not to maximize acquired information, since anchoring is complete once every ambiguity capable of changing the interaction-relevant quantity has been resolved.

A local differential characterization gives a quantitative lower bound.
Let us assume that $\mathcal Q_I$ is a smooth manifold and that $C_I$ and $\Theta_I$ are differentiable near $q$.
Let us define an \textit{epistemically invisible} tangent space $V_C(q)=\ker DC_I(q)$ and the interaction-invariant tangent space $V_\Theta(q)=\ker D\Theta_I(q)$.
Directions in $V_C(q)$ are undetectable to first order from relational evidence, whereas directions in $V_\Theta(q)$ do not affect the interaction-relevant quantity to first order.
Their mismatch is
\[
r_\Theta(q)
=
\dim V_C(q)-\dim\big(V_C(q)\cap V_\Theta(q)\big)
=
\operatorname{rank}\!\left(D\Theta_I(q)\big|_{V_C(q)}\right).
\]
Thus $r_\Theta(q)$ counts the independent local directions that are invisible to relational evidence but visible to the interaction.

\noindent\textit{Theorem 1 (Local anchor-dimension lower bound).}
Let $\zeta:\mathcal Q_I\rightarrow\mathbb R^m$ be a bundle of $m$ scalar admissible anchors.
If the augmented evidence is first-order $\Theta$-sufficient at $q$, in the sense that
\[
\ker D(C_I,\zeta)_q\subseteq V_\Theta(q),
\]
then
\[
m\geq r_\Theta(q).
\]

\noindent\textit{Proof.}
Let $U=V_C(q)$ and $W=V_C(q)\cap V_\Theta(q)$.
First-order sufficiency gives
\[
\ker D(C_I,\zeta)_q=U\cap\ker D\zeta_q,
\qquad
\ker(D\zeta_q|_U)\subseteq W.
\]
Hence
\[
\operatorname{rank}(D\zeta_q|_U)
=
\dim U-\dim\ker(D\zeta_q|_U)
\geq
\dim U-\dim W
=
r_\Theta(q).
\]
Since the rank of a map into $\mathbb R^m$ cannot exceed $m$, the result follows.
\hfill$\square$

The bound does not guarantee that suitable physical anchors exist.
Equality requires the embodied observer agents' sensorimotor interface to provide anchors that actually resolve the required directions.

\noindent\textit{Corollary 3 (Equality under anchor richness).}
Let us suppose that the admissible anchor family contains $r_\Theta(q)$ scalar anchors whose bundle $\zeta$ satisfies
\[
\ker\!\left(D\zeta_q\big|_{V_C(q)}\right)
=
V_C(q)\cap V_\Theta(q).
\]
Then the minimum local number of scalar anchors satisfies $m_{\min}^{\mathrm{local}}=r_\Theta(q)$.
We refer to this as the \emph{anchor richness condition}: the information-theoretically required mismatch is also resolvable through admissible sensorimotor coupling.

\noindent\textit{Running example.}
For $C_I(q_A,q_B)=q_A-q_B$, we have $DC_I=[1,-1]$ and therefore $V_C=\operatorname{span}\{(1,1)\}$, that is, the invisible local direction is the common shift.
For \textsf{S1-R}, the relative response has the same invariant direction, so $V_{\Theta_R}=V_C$ and $r_{\Theta_R}=0$.
For \textsf{S1-A}, an absolute response such as $\Theta_A(q)=(q_A-t_H,q_B-t_H)$ has $V_{\Theta_A}=\{0\}$, hence $r_{\Theta_A}=1$.
The lower bound therefore already predicts the qualitative result developed explicitly in Section~\ref{sec:toyworld}, that is, zero additional scalar anchors for the relative scenario and at least one for the absolute scenario.

The results of this Section can be summarized by the same stopping condition introduced in Section~\ref{sec:model}.
After any necessary sensorimotor anchoring, it is sufficient that $\Omega_{C,K}(q)\subseteq[q]_\Theta$.
The next Section applies these results to the two asynchronous hand-off scenarios \textsf{S1-R} and \textsf{S1-A}, which share the same observer agents, representations, transport maps, and relational evidence and differ only in the invariance structure of the interaction.

\section{Two Variants of the Asynchronous Hand-off Scenario}
\label{sec:toyworld}

In this Section, we instantiate the theory developed in Section~\ref{sec:theory} in the asynchronous hand-off scenario introduced in Section~\ref{sec:background}. 
The two variants, namely \textsf{S1-R} and \textsf{S1-A}, share the same physical process, observer agents, native representations, representational transports, and relational evidence.
They differ only in what the interaction treats as relevant. 
This controlled contrast isolates interaction invariance as the factor determining whether additional sensorimotor anchoring is necessary.

\subsection{Common observer model and unresolved common-mode ambiguity}

Let us consider an object moving in one dimension according to
\[
x(t) = x_0 + vt, \qquad v > 0,
\]
toward a hand-off location $x_H$. 
The corresponding physical hand-off time is $t_H = (x_H - x_0)/v$. 
The quantity $t_H$ is not assumed to be directly represented or available to either observer agent.

Agent $A$ observes the process at time $t_A$ and represents the future hand-off through estimates of remaining distance and velocity, such that $R_A = (\hat d_A, \hat v_A)$. 
If we write $\tau_A = t_H - t_A$, $d_A = v\tau_A$, $\hat d_A = d_A + \varepsilon_d$, and $\hat v_A = v+\varepsilon_v$, its transport into the hand-off-time comparison context is
\[
q_A
=
\Phi_A^{c_H}(R_A)
=
t_A+\frac{\hat d_A}{\hat v_A}.
\]
Agent $B$ observes at a generally different time $t_B$ and uses the heterogeneous native representation $R_B = \widehat{\tau}_B$, where
$\widehat{\tau}_B = (t_H - t_B) + \varepsilon_\tau$. 
Its corresponding transport is
\[
q_B
=
\Phi_B^{c_H}(R_B)
=
t_B+\widehat{\tau}_B.
\]
As a consequence, we have $\mathcal R_A \neq \mathcal R_B$, while both transported contents lie in the same interaction-specific comparison space.

Let $e_A = q_A - t_H$ and $e_B = q_B - t_H$ denote the transport errors. 
Direct substitution gives
\[
e_A
=
\frac{\varepsilon_d-\tau_A\varepsilon_v}{v+\varepsilon_v},
\qquad
e_B = \varepsilon_\tau.
\]
The expression for $e_A$ makes explicit that asynchronous observation and prediction horizon affect transport accuracy, but no particular noise model is
required for the argument below.

As anticipated in the running example of Section~\ref{sec:theory}, let us set $q = (q_A, q_B)$ and, suppressing the interaction index for readability, let us define the relational evidence as
\[
C(q) = q_A - q_B = e_A - e_B.
\]
A simple agreement test is $|C(q)| \leq \epsilon$. 
The complete ambiguity fiber is
\[
\Omega_C(q)
=
\left\{
(q_A + \delta, q_B + \delta):
\delta\in\mathbb R
\right\}.
\]
As a consequence, the pairwise evidence is blind to the common-mode direction $(1,1)$, that is, all transformations $g_\delta(q_A, q_B) = (q_A + \delta, q_B + \delta)$ leave $C$ unchanged.

\begin{figure}[t!]
\centering
\begin{tikzpicture}[scale=1.08]
    \draw[->] (-2.3,0) -- (2.5,0) node[right] {$e_A$};
    \draw[->] (0,-2.3) -- (0,2.5) node[above] {$e_B$};

    \fill[gray!10] (-1.0,-1.0) rectangle (1.0,1.0);
    \draw[thick] (-1.0,-1.0) rectangle (1.0,1.0);
    \node[anchor=north west] at (-0.92,0.92)
    {\scriptsize adequate};

    \draw[dashed,thick]
        (-2.0,-2.2) -- (2.2,2.0)
        node[pos=0.13,sloped,below=5pt]
        {\small $C(q)=C(q')$};

    \coordinate (q)  at (0.35,0.15);
    \coordinate (qp) at (1.70,1.50);

    \fill (q)  circle (1.8pt);
    \fill (qp) circle (1.8pt);

    \node[anchor=north west,
          xshift=4pt,yshift=-2pt]
          at (q) {\small $q$};

\node[anchor=west,
      xshift=10pt,
      yshift=8pt]
      at (qp) {\small $q'$};
      
    \draw[->,very thick,
          shorten >=3pt,
          shorten <=3pt]
          (q) -- (qp)
          node[pos=0.58,
               above left,
               xshift=-2pt,
               yshift=3pt]
          {$g_\delta$};
\end{tikzpicture}
\caption{
A geometrical depiction of the asynchronous hand-off scenario in transport-error coordinates. 
The shaded region represents the absolute-timing tolerances of \textsf{S1-A}. 
The configuration $q$ lies inside that region, while $q' = g_\delta q$ lies outside it. 
Both belong to the same relational-evidence fiber because $e_A - e_B$ is invariant to common-mode translation. 
The same translation is interaction-invariant in \textsf{S1-R} but interaction-relevant in \textsf{S1-A}.}
\label{fig:tw1_geometry}
\end{figure}
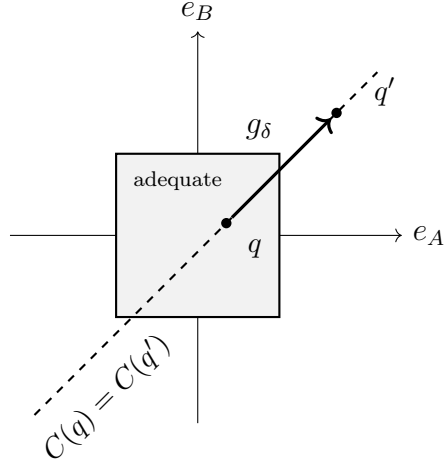

This common-mode ambiguity should not be confused with a change of temporal
coordinates. 
Shifting $(q_A, q_B, t_H)$ jointly by $\delta$ changes only the temporal origin, whereas shifting $(q_A, q_B)$ while keeping $t_H$ fixed changes both transported representations relative to the physical event.
Only the latter transformation is relevant to the sufficiency analysis.
Figure~\ref{fig:tw1_geometry} summarizes this geometry in error coordinates.

For this analytical example we use the simplest fixed action policy: each observer schedules its task action at its transported event-time estimate, so that $T_A=q_A$ and $T_B=q_B$. 
The response maps below are therefore functions of $q$, consistently with the fixed-protocol assumption of Section~\ref{sec:model}.

\subsection{\textsf{S1-R}: relative synchronization}

In \textsf{S1-R}, the interaction requires only relative synchronization between the two observer agents.
The response and adequacy maps are
\[
\Psi_R(q)=q_A-q_B,
\qquad
\mathcal A_R(q)
=
\mathbf 1\!\left[|q_A-q_B|\leq\eta_{AB}\right].
\]
Since $\Psi_R = C$, both the response and adequacy maps factor through the relational evidence. 
Proposition~1 therefore implies that $C$ is both response-sufficient and adequacy-sufficient.
The geometric reason is that
\[
\Omega_C(q)
=
[q]_{\Psi_R}
=
\left\{
(q_A+\delta,q_B+\delta):
\delta\in\mathbb R
\right\}.
\]
The common-mode transformations are invisible not only to $C$ but also to the interaction. 
There are therefore no conflicting epistemic pairs along this ambiguity, and Proposition~2 requires no additional anchor.
The local dimension result reaches the same conclusion. 
Here
$DC = D\Psi_R = [1,-1]$, so
\[
V_C
=
V_{\Psi_R}
=
\operatorname{span}\{(1,1)\},
\qquad
r_{\Psi_R}=0.
\]
As a consequence, the common shift is epistemically invisible but also interaction-invariant, whereas the minimum number of additional scalar anchors is
\[
m_{\min}^{R} = 0.
\]
Therefore, \textsf{S1-R} is a limiting case in which relational evidence already preserves exactly the distinctions needed for the interaction.
Resolving the absolute temporal offset would \textit{refine} the representation without changing the interaction-relevant quantity.

\subsection{\textsf{S1-A}: absolute hand-off timing and minimal anchoring}

\textsf{S1-A} retains the same observer model, transports, and relational evidence, but the interaction now distinguishes timing relative to the
physical hand-off event. 
Its response signature is
\[
\Psi_A(q)
=
\begin{bmatrix}
q_A - t_H\\
q_B - t_H
\end{bmatrix}
=
\begin{bmatrix}
e_A\\
e_B
\end{bmatrix},
\]
and a representative adequacy predicate is
\[
\mathcal A_A(q)
=
\mathbf 1
\left[
|e_A|\leq\eta_A
\;\land\;
|e_B|\leq\eta_B
\right].
\]
A separate pairwise synchronization tolerance could be added without changing the main argument.

A common-mode shift leaves relational evidence unchanged but changes the response, that is,
\[
C(g_\delta q) = C(q),
\qquad
\Psi_A(g_\delta q)
=
\Psi_A(q)
+
\delta
\begin{bmatrix}
1\\
1
\end{bmatrix}.
\]
For every non-zero $\delta$, the two configurations have the same relational evidence but different response values. 
Hence, it holds that $\Omega_C(q) \nsubseteq [q]_{\Psi_A}$, and by Corollary~1, $C$ is not response-sufficient for \textsf{S1-A}.
The same failure can occur at the adequacy level. 
For example, let us assume $e_A = e_B = 0$. 
A sufficiently large common shift produces $e_A' = e_B' = \delta$.
In this case, agreement remains perfect because $C(q') = C(q) = 0$, but the absolute timing tolerances are violated. 
Therefore, identical relational evidence can correspond to both adequate and inadequate interaction.

To remove this common-mode ambiguity, let us consider an admissible sensorimotor \textit{phase} anchor. 
Let the moving object cross an observable landmark $L$ before hand-off at
\[
t_L = t_H - \Delta_L,
\qquad
\Delta_L>0,
\]
where $\Delta_L$ is a known interaction-specific offset in the idealized scenario. 
Detecting the landmark yields the locally obtainable event-time estimate $\widehat t_H^{\,L} = t_L + \Delta_L$, which equals $t_H$ in the ideal
model. 
Define the scalar anchor
\[
\zeta_L(q)
=
q_A-\widehat t_H^{\,L}
=
e_A.
\]
This anchor does not contribute to a global world representation. 
On the contrary, it supplies one sensorimotor relation between the observer $A$'s transported representation and an observable environmental event.
If we write $d = C(q) = q_A - q_B$ and $a = \zeta_L(q) = e_A$, we obtain
\[
e_B = a-d,
\qquad
\Psi_A(q)
=
\begin{bmatrix}
a\\
a-d
\end{bmatrix}.
\]
As a consequence, the augmented evidence $(C, \zeta_L)$ is response-sufficient. 
However, in this case, zero anchors cannot be sufficient because the common-mode shift gives conflicting epistemic pairs, whereas the single scalar anchor $\zeta_L$ separates that ambiguity.
Also in this case, the differential characterization gives the same result. 
Since $V_C = \operatorname{span}\{(1,1)\}$ and $D\Psi_A = I_2$, we have $V_{\Psi_A} = \{0\}$ and therefore $r_{\Psi_A} = 1$. 
Theorem~1 imposes $m \geq 1$. 
Moreover, $D\zeta_L = [1, 0]$ acts non-trivially on the invisible common-mode direction, so the anchor-richness condition is satisfied. 
Consequently, we have
\[
m_{\min}^{A}=1.
\]

The controlled \textit{contrast} between the two scenarios is therefore
\[
m_{\min}^{R}=0,
\qquad
m_{\min}^{A}=1.
\]
Nothing about sensing accuracy, native representation quality, or transport
mechanism changes between the two cases. 
The difference is induced entirely by the mismatch between the invariances of the relational evidence and those of the interaction.

\noindent\textit{Adequacy can require less information.}
Exact reconstruction of the response $\Psi_A$ is not always necessary. 
Let us suppose that an admissible sensorimotor cue reports only whether observer $A$ lies within an absolute timing window, such that
\[
Z_\rho
=
\mathbf 1\!\left[|e_A|\leq\rho\right].
\]
If $Z_\rho=1$ and the relational evidence guarantees $|e_A - e_B| \leq \epsilon$, then we have $|e_B| \leq \rho+\epsilon$. 
Therefore
\[
\rho\leq\eta_A,
\qquad
\rho+\epsilon\leq\eta_B
\]
are sufficient to guarantee $\mathcal A_A(q)=1$.
The Boolean cue does not reconstruct $(e_A, e_B)$ and is therefore not response-sufficient, but it is sufficient to establish adequacy over the
specified tolerance regime. 
In information terms, full representational identification is therefore stronger than response sufficiency, which is in
turn stronger than adequacy sufficiency. 
The two scenarios are summarized and compared in Table~\ref{tab:tw1_comparison}.

\begin{table}[t!]
\centering
\scriptsize
\caption{
Controlled comparison of the two asynchronous hand-off scenarios.
The observer model and relational evidence are identical; 
only the interaction invariance changes.}
\label{tab:tw1_comparison}
\begin{tabular}{lcc}
\toprule
 & \textsf{S1-R} & \textsf{S1-A} \\
\midrule
Observer agents & same & same \\
Native representations & same & same \\
Transport maps & same & same \\
Relational evidence $C = q_A - q_B$ & same & same \\
Common-mode ambiguity & present & present \\
Common mode interaction-invariant & yes & no \\
$C$ response-sufficient & yes & no \\
Additional scalar anchor for exact response & $0$ & $1$ \\
$r_{\Psi}$ & $0$ & $1$ \\
Unique joint configuration required & no & no \\
\bottomrule
\end{tabular}
\end{table}

The pair \textsf{S1-R}/\textsf{S1-A} provides a controlled theoretical test of relational sufficiency. 
In \textsf{S1-R}, the ambiguity preserved by relational evidence lies entirely inside the interaction-equivalence class and should remain unresolved. 
In \textsf{S1-A}, the same ambiguity crosses interaction-equivalence classes and must be reduced by one independent sensorimotor relation to the environment.

\section{Discussion}
\label{sec:discussion}

The results above support a specific claim rather than a general rejection of representational alignment. 
Heterogeneous embodied observer agents sometimes need additional representational compatibility in order to interact coherently. 
The question we focus on in this paper is how much compatibility the interaction actually requires. 
The analysis of Sections~\ref{sec:theory} and \ref{sec:toyworld} suggests that this requirement is determined by the relation between two structures, namely the ambiguities preserved by the available \textit{relational evidence} and the distinctions induced by the \textit{interaction-relevant map}.

This perspective shifts the objective from reconstructing a \textit{uniquely aligned} configuration to identifying a \textit{sufficient} relational quotient. 
It also gives sensorimotor interaction a specific theoretical role. 
The environment need not allow for a globally accessible representation against which local, observer-specific representations are compared. 
Instead, environmental interaction can supply selective constraints that eliminate only those ambiguities that matter to the task.

\subsection{Alignment, anchoring, and interaction-relevant representation}

A central consequence of the theory is that representational disagreement is not intrinsically problematic. 
A representational difference matters only when it crosses interaction-equivalence classes.
Let $L(q)$ denote any aligned representation sufficient for an interaction-relevant quantity $\Theta_I$. 
If $q\sim_\Theta q'$ but $L(q)\neq L(q')$, then $L$ distinguishes configurations that cannot alter the chosen interaction-relevant quantity.
In this case, the alignment is task-over-refined in the sense introduced in Section~\ref{sec:theory}. 
This does not imply that retaining the distinction is detrimental to an observer's more general knowledge. 
It may be useful for another task, for transfer, or for future interactions. 
It only establishes that the distinction is not required for the interaction currently under consideration.

This observation is important when interpreting the relationship between relational sufficiency and approaches based on common representations,
relative representations, or local alignment. 
The theory developed in this paper does not replace those mechanisms. 
Representational transport $\Phi_\alpha^c$ may itself be implemented through learned alignment, coordinate transformation, relative encoding, prediction, semantic translation, or another comparison mechanism. 
Instead, our claim concerns the stopping condition: once Equation~\eqref{eq:relational_sufficiency_condition} holds, a further reduction of representational ambiguity is not required for $\Theta$-sufficiency.

The variant \textsf{S1-R} gives the simplest example. 
The common-mode degree of freedom is unobservable from relational evidence but is also irrelevant to the relative task. 
Resolving it would therefore improve identifiability without improving task-relevant identifiability. 
\textsf{S1-A} differs because the physical interaction distinguishes the same degree of freedom, and a sensorimotor anchor is therefore required.

This interpretation also clarifies the role of anchoring. 
Anchoring should not be understood as reintroducing a partial, privileged world representation. 
The landmark used in \textsf{S1-A}, for example, does not reveal a complete environment state. 
On the contrary, it establishes a single additional relation between an observer-relative estimate and an environmental event. 
That relation is sufficient because the unresolved ambiguity has only one task-relevant dimension.
More generally, the local quantity
\[
r_\Theta(q)
=
\dim V_C(q)
-
\dim
\left(
V_C(q)\cap V_\Theta(q)
\right)
\]
can be interpreted as a mismatch between the epistemic symmetries of the observers and the invariances of the task. 
Importantly, it does not measure \textit{how much} the agents fail to represent the world;
instead, it measures how many locally independent distinctions are simultaneously invisible to their relational evidence, and visible to the interaction-relevant quantity.

As discussed in Section~\ref{sec:background}, this point is deliberately close to, but distinct from, functional observability. Functional observability asks whether a target function of a system state can be reconstructed without reconstructing the complete state, and recent work also optimizes sensor sets for that purpose \citep{Montanari2022PRR, Montanari2022PNAS, Zhang2025}. 
Therefore, Theorem~1 should not be read as a new observability principle. 
Its role is to quantify the \emph{relational} deficit left after heterogeneous observer-relative representations have already been transported into a
comparison context, with admissible sensorimotor anchors restricted by the agents' embodied coupling.

The distinction between response sufficiency and adequacy sufficiency strengthens this interpretation. Reconstructing $\Psi_I(q)$ may require more information than determining $\mathcal A_I(q)$. 
The Boolean anchor of Section~\ref{sec:toyworld} illustrates this point: bounded relational evidence together with a one-bit environmental constraint may guarantee successful interaction even though the \textit{exact physical response} remains unidentified. 
There is therefore a hierarchy between complete representational identification, task-response identification, and the
information required merely for successful action.

This hierarchy also suggests a natural implication for learning systems. 
For example, in a reinforcement-learning formulation, an agent receives task-dependent reward rather than supervision requiring reconstruction of every latent variable.
If representational or communication complexity carries a cost, an optimal solution may have no incentive to resolve differences lying entirely inside
an interaction-equivalence class. 
In this sense, reward-driven learning may favour representations of $[q]_\Theta$ rather than globally aligned representations of $q$.
This is not presented here as a new result in reinforcement learning.
Task-relevant state abstraction, bisimulation-based aggregation, and learned compact state models are already mature ideas \citep{Ferns2004, Abel2016, Gelada2019}. 
The present implication is narrower, that is, in a multi-observer embodied setting, the object that may be quotiented is the
\emph{joint relational configuration} of heterogeneous representations. 
The resulting question for learning becomes which discrepancies between observer-relative representations need to be resolved for the reward-relevant interaction.
Empirical investigation of this hypothesis is left for future work.

The same qualification applies to decentralized sufficient-information and goal-oriented communication frameworks. 
They already demonstrate that agents need not preserve or exchange all available information \citep{Nayyar2013, Tavafoghi2018, Charalambous2026}. 
The present contribution is to use interaction relevance to define a stopping condition on \emph{inter-representational resolution}, that is, which differences between embodied observers must become distinguishable, and which may remain unresolved.

\subsection{What changes with one or more than two observers?}

This paper deliberately treats interactions between two distinct observer agents, which is the smallest setting in which genuinely inter-observer relational ambiguity can appear.
This setting is sufficient to expose the mismatch between relational evidence and task invariance without introducing additional structures.

One may ask what remains if the assumption of two \textit{distinct} observers is dropped. 
The formulation of the theory makes the degenerate case straightforward. 
Let us assume that $i=j$, while for the events we have $\alpha\neq\beta$.
The two events may correspond to different times, viewpoints, modalities, sensorimotor histories, or representational contexts of the \textit{same} embodied agent. 
Representational transport and relational evidence can then be defined exactly as before. 
The framework becomes a theory of relational sufficiency
between different \textit{situated} representations of a single observer.
For example, an agent acting at time $t$ may compare a present representation with a predicted representation at $t + \Delta t$, or compare two views acquired from different locations. 
A residual ambiguity may again be \textit{harmless} for one future action and \textit{consequential} for another. 
Formally, the distinction between inter-observer and intra-observer comparison is therefore not fundamental.
However, the single-observer case is related to different scientific questions.
It approaches matters related to state abstraction, prediction, sensorimotor invariance, and temporal self-consistency that are not the main focus of the present paper. 
We consequently treat $i=j$ as a degenerate extension rather than as part of the main contribution.

The opposite extension, that is, having more than two observer agents, is potentially more consequential. 
Pairwise relational sufficiency need not imply sufficiency of the complete multi-observer \textit{system}. 
With three or more observers, additional structures become possible, for example, pairwise relations may be individually acceptable yet mutually inconsistent, relational congruity may be non-transitive, and different subsets of observers may require different comparison contexts or anchors.
A natural representation could be that of a graph or hypergraph whose nodes are representation events and whose edges or hyperedges encode comparison relations, transports, or interaction constraints. 
Cycles over graphs may introduce a new question absent from the two-observer analysis, that is, whether a sequence of locally admissible transports returns a representation to a configuration compatible with its starting point. 
Likewise, an anchor acquired by one observer may propagate through some parts of the relational structure but not others.

The theory developed here suggests, but does not establish, a multi-observer generalization in which one may ask whether the residual ambiguity of an entire relational network lies inside the equivalence class induced by a collective interaction. 
Such a theory would need to address cycle inconsistency, non-transitive congruity, higher-order constraints, and the distribution of anchors across the network. 

\subsection{Scope, assumptions, and limitations}

Several restrictions of the current analysis are defined on purpose.

\textit{First}, the main results are stated with respect to a fixed interaction and a fixed interaction-relevant map.
Interaction equivalence is therefore conditional on $I$ and on the chosen $\Theta_I$. 
A distinction that is irrelevant for one task may become essential for another. 
Relational sufficiency should not be interpreted as a claim that the discarded distinctions are globally meaningless, but that they need not be resolved for the specified interaction.

This dependence is especially important when agents must support many interactions over long time horizons. If $\mathcal I = \{I_1, \ldots, I_m\}$ is a family of possible interactions, the relevant equivalence relation would need to preserve all task responses of interest, potentially refining the quotient as the task family grows. 
In this paper, we consider one interaction at a time and we do not address how a persistent general-purpose representation should be selected across a changing task distribution.

\textit{Second}, the theory assumes that a suitable representational transport has already been defined. 
The transport maps $\Phi_\alpha^c$ may be imperfect, learned, context-dependent, or even unavailable in practice.
This paper investigates what relational information is sufficient \emph{after} comparison has been made possible. 
It does not solve the separate problem of learning or constructing the transport operators themselves.

\textit{Third}, the anchors analysed here are intentionally simple. 
They are assumed to be admissible through the agents' sensorimotor capabilities and available before completion of the relevant task action. 
In \textsf{S1-A}, observing a landmark provides information without materially altering the underlying task. 
In general, embodied agents can actively create anchors by moving, touching, waiting, querying another agent, or performing an experiment designed to disambiguate competing representations. 
In such cases the probing action may change both the epistemic configuration and the physical state. 
Extending the current static anchoring results to active, state-changing information acquisition is an important next step.

\textit{Fourth}, the deterministic task-response map $\Psi_I(q)$ is an idealization. 
A stochastic interaction can instead be characterized by a conditional response distribution $P_I(\cdot\mid q)$, with interaction equivalence defined by equality of the relevant distributions, or by equality with respect to a weaker decision criterion. 
The factorization logic of relational sufficiency extends naturally to this setting, but the current paper retains the deterministic form to expose the structural result without additional probabilistic machinery.

\textit{Fifth}, Theorem~1 is local and first-order. 
The quantity $r_\Theta(q)$ characterizes differential mismatch near a configuration, but it does not by itself establish global sufficiency in nonlinear spaces. 
Likewise, equality between the lower bound and the number of required anchors depends on the anchor-richness condition. 
A system may theoretically require one additional degree of information while lacking any physically available sensorimotor measurement capable of providing it.

Finally, the scenario considered here is deliberately analytically solvable rather than being an embodied benchmark. 
Its purpose is to distinguish two theoretical possibilities under controlled conditions. 
The contrast $m_{\min}^{R}=0$ versus $m_{\min}^{A}=1$ is informative because the observer model and relational evidence are held fixed while the task changes. 
Physical or simulated robot experiments would address a different question, that is, whether the same principles remain useful under imperfect transport, uncontrolled sensing, actuation uncertainty, learned representations, communication failures, and richer action spaces.
The absence of such experiments therefore limits the present paper's claims about external validity, but not its intended theoretical claim. 

\section{Conclusions}
\label{sec:conclusion}

This paper investigates which relational distinctions between observer-relative representations must be resolved for coherent embodied interaction, and which may remain unresolved because the interaction itself cannot distinguish them.

We considered embodied observer agents with heterogeneous local representations that need not be directly comparable and need not converge to
a common world representation. 
Representational transport constructs interaction-specific comparison quantities, while relational evidence captures what can be established among them. 
The central result is that such evidence need not identify a unique joint representational configuration. 
Specifically, relational evidence is sufficient whenever its residual ambiguity lies entirely within an equivalence class induced by the interaction, that is, when $\Omega_{C,K}(q)\subseteq[q]_\Theta$.
If this condition fails, additional sensorimotor anchoring is required only along those representational degrees of freedom that are simultaneously
invisible to relational evidence and relevant to the interaction.

These considerations are illustrated by two variants of the same analytical hand-off scenario.
The relative-synchronization variant requires no additional scalar anchor, whereas the absolute-timing variant requires one for exact response sufficiency; the difference arises solely from the interaction invariances.

The results suggest a different stopping condition for representational alignment in embodied cognitive systems. 
The objective may not be \textit{complete agreement}, full state identification, or convergence toward a privileged shared representation. 
Instead, alignment or grounding need only proceed until the ambiguity that remains is \textit{behaviourally irrelevant} to the interaction.

The present analysis establishes this principle for two embodied observer agents and analytically controlled tasks. 
Extending it to active state-changing anchoring, stochastic interactions, learned representational transport, and networks of multiple observers provides the next step toward a broader theory of relational consistency in embodied cognitive systems.

\bibliographystyle{elsarticle-harv}
\bibliography{references}

\end{document}